\documentclass[conference]{IEEEtran}

\IEEEoverridecommandlockouts
\usepackage{cite}
\usepackage{amsmath,amssymb,amsfonts}
\usepackage{bbm}
\usepackage{algorithmic}
\usepackage{graphicx}
\usepackage{textcomp}
\usepackage{xcolor}
\usepackage{url}
\usepackage{booktabs}
\usepackage{placeins}
\usepackage{makecell} 
\usepackage{caption}
\usepackage{subcaption}
\usepackage{pifont}
\usepackage{flushend}
\usepackage{hyperref}

\usepackage[T1]{fontenc}

\begin{document}

\title{Investigating Location-Regularised Self-Supervised Feature Learning for Seafloor Visual Imagery\\
}

\author{
\IEEEauthorblockN{Cailei Liang}
\IEEEauthorblockA{
  \textit{University of Southampton}\\
  Southampton, UK \\
  c.liang@soton.ac.uk
}
\and
\IEEEauthorblockN{Adrian Bodenmann}
\IEEEauthorblockA{
  \textit{University of Southampton}\\
  Southampton, UK \\
  adrian.bodenmann@soton.ac.uk
}
\and
\IEEEauthorblockN{Emma J Curtis}
\IEEEauthorblockA{
  \textit{University of Southampton}\\
  Southampton, UK \\
  e.j.curtis@soton.ac.uk
}
\and
\IEEEauthorblockN{Samuel Simmons}
\IEEEauthorblockA{
  \textit{University of Southampton}\\
  Southampton, UK \\
  s.j.simmons@soton.ac.uk
}
\and
\IEEEauthorblockN{Kazunori Nagano}
\IEEEauthorblockA{
  \textit{IIS, The University of Tokyo}\\
  Tokyo, Japan \\
  k-nagano@iis.u-tokyo.ac.jp
}
\and
\IEEEauthorblockN{Stan Brown}
\IEEEauthorblockA{
  \textit{Voyis Imaging Inc.} \\
  Ontario, Canada \\
  stan@voyis.com
}
\and
\IEEEauthorblockN{Adam Riese}
\IEEEauthorblockA{
  \textit{Voyis Imaging Inc.} \\
  Ontario, Canada \\
  adam@voyis.com
}
\and
\IEEEauthorblockN{Blair Thornton}
\IEEEauthorblockA{
  \textit{University of Southampton, UK}\\
  \textit{IIS, The University of Tokyo, Japan}\\
  b.thornton@soton.ac.uk
}
}

\maketitle

\begin{abstract}
High-throughput interpretation of robotically gathered seafloor visual imagery can increase the efficiency of marine monitoring and exploration. Although recent research has suggested that location metadata can enhance self-supervised feature learning (SSL), its benefits across different SSL strategies, models and seafloor image datasets are underexplored. This study evaluates the impact of location-based regularisation on six state-of-the-art SSL frameworks, which include Convolutional Neural Network (CNN) and Vision Transformer (ViT) models with varying latent-space dimensionality. Evaluation across three diverse seafloor image datasets finds that location-regularisation consistently improves downstream classification performance over standard SSL, with average F1-score gains of 4.9$\pm$4.0\% for CNNs and 6.3$\pm$8.9\% for ViTs, respectively. While CNNs pre-trained on generic datasets benefit from high-dimensional latent representations, dataset-optimised SSL achieves similar performance across the high (512) and low (128) dimensional latent representations. Location-regularised SSL improves CNN performance over pre-trained models by 2.7$\pm$2.7\% and 10.1$\pm$9.4\% for high and low-dimensional latent representations, respectively. The best-performing CNN was a low-dimensional location-regularised SSL. This achieved an F1-score of 0.778$\pm$0.115, enhancing the best pre-trained CNN score by 6.0$\pm$1.4\%. For ViTs, high-dimensionality benefits both pre-trained and dataset-optimised SSL. Although location-regularisation improves SSL performance compared to standard SSL methods, pre-trained ViTs show strong generalisation, matching the best-performing location-regularised SSL with F1-scores of 0.795$\pm$0.075 and 0.795$\pm$0.077, respectively. The findings highlight the value of location metadata for SSL regularisation, particularly when using low-dimensional latent representations, and demonstrate strong generalisation of high-dimensional ViTs for seafloor image analysis.
\end{abstract}


\section{Introduction}
Visual images are routinely used in seafloor monitoring and exploration due to the high resolution and flexibility of information they provide~\cite{williams}. The large volume of images collected by autonomous underwater vehicles (AUVs) motivates the use of machine learning to enhance processing efficiency. Self-supervised learning (SSL) allows latent representations to be extracted from images without the need for human-annotated datasets, which require time and effort to produce~\cite{jaiswal2020survey}. These can be used as features for downstream tasks such as unsupervised clustering, content based-query, as well as semi-supervised training of downstream classifiers, which require significantly fewer human-labelled examples compared to fully supervised approaches~\cite{simclr, yamada-lga}. 

\begin{figure}
    \centering
    \includegraphics[width=1\linewidth]{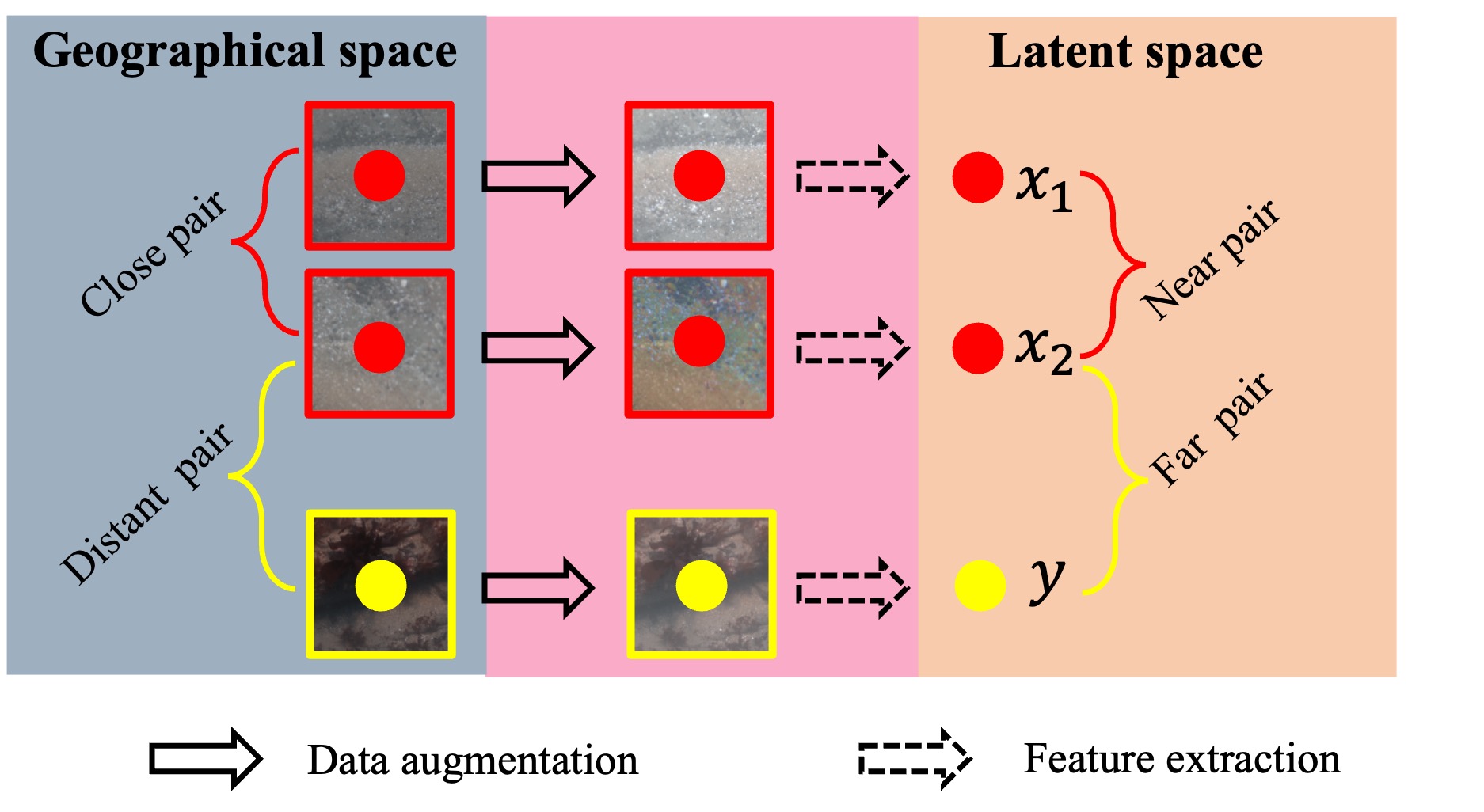}
    \caption{Location regularisation in contrastive SSL. Instead of generating a positive pair (red) using augmented views of the same image, physically nearby images are used. It is assumed that they share similar attributes, but capture natural variability within the same semantic group. These are contrasted with negative pairs (yellow) formed from distant images~\cite{yamada-geoclr}.}
    \label{fig:location metadata}
\end{figure}

 Various SSL strategies have been developed to discriminate between patterns in imagery without the need for labelled training data~\cite{simclr, moco}. These typically use data augmentation methods~\cite{vonkugelgen2021ssl} and train models such as CNNs (Convolutional Neural Networks) or ViT (Vision Transformers)~\cite{simclr,dino}. Most state-of-the-art approaches rely on the concept of positive views, where models leverage an assumed similarity of appearance between augmented views generated from the same image. They do this by embedding these views to nearby regions of a latent representation space. When imaging the seafloor, camera-equipped robots use strobes and operate at low altitudes to overcome the strong attenuation of light in water. The resulting images cover only a small extent of seafloor (1-10\,m). Incorporating location metadata can benefit seafloor image interpretation~\cite{yamada-pami} because the substrates and habitats that characterise appearance typically vary at spatial scales that are much larger than the extent of a single image frame~\cite{zelada}. Latent representations generated by sampling positive views from physically nearby images (see Fig.~\ref{fig:location metadata} instead of the same image have demonstrated enhanced performance across a number of downstream tasks as this allows them to avoid overfitting the natural variability that exists within semantic groupings~\cite{yamada-geoclr}. This study aims to understand whether the benefits of leveraging location metadata generalises across different SSL training methods. We investigate the impact of location-regularisation beyond contrastive CNNs~\cite{yamada-geoclr}, investigate its impact on multi-view CNNs and ViT models, and further investigate the impact of latent space dimensionality across different methods. Our contributions are:
\begin{itemize}
    \item We investigate the effect of location-based regularisation in five state-of-the-art CNN-based SSL approaches; SimCLR, SimSiam, MoCo-v2, SwAV, and DeepCluster-v2~\cite{simclr,simsiam,moco,swav,deepcluster}).
    \item We extend location-based regularisation to a state-of-the-art ViT-based model (DINO~\cite{dino}) and investigate its effect on SSL-based ViT fine-tuning for optimisation to each target dataset.
    \item We evaluate the impact of using different latent-representation dimensionalities on pre-trained, SSL and location-regularised SSL trained models.
\end{itemize}
\noindent We evaluate each method's performance using downstream classifiers that delineate their latent representations. The analysis is performed on datasets that were gathered under various conditions using different AUV imaging systems across a range of habitats. These datasets consist of over 90k images with human-validated reference labels. Section II reviews the literature review on state-of-the-art SSL approaches; Section III outlines the key principles of various discriminative SSL methods, which we adapt to leverage location metadata for regularisation. Section IV evaluates the performance of the models, with conclusions in Section V.

\section{Literature Review}

State-of-the-art SSL methods can be broadly categorised into context-based, generative and discriminative approaches~\cite{gui2024survey}. Context-based SSL aims to capture structural relationships within data and is often used for reasoning in natural language processing (NLP)~\cite{devlin-etal-2019-bert}. Generative SSL methods typically focus on reconstructing degraded inputs, restoring masked elements, or making predictions to construct new data~\cite{openai2024gpt4technicalreport}. While they are effective at capturing fine-grained structural details, discriminative SSL methods are typically more effective at capturing distinctive patterns that can be used to differentiate between different semantic groupings~\cite{liu2023ssl}.

Discriminative SSL approaches can be split into instance-level, cluster-level, and self-distillation methods~\cite{liu2023ssl}. Instance-level methods typically use image augmentation methods such as flipping, cropping, blurring, distortion and colour modification to generate two or more augmented views of the same input. The augmented views are passed through an encoder to obtain their latent vectors, or representations, where model parameters are adjusted to minimise a loss function that is designed to pull positive views together in the model's latent space. To avoid trivial solutions where the encoders collapse all inputs to the same latent vector, contrastive learners such as SimCLR introduce negative views, generated from different images, that get pushed apart in the latent space using a contrastive loss function~\cite{simclr}. During optimisation, SimCLR uses large batches of positive and negative pairs that are simultaneously processed to learn CNN parameters that generalise. A key disadvantage of this approach is the high computational and memory cost associated with processing and storing negative pairs, which in turn limits the amount of informative signal per batch (i.e., the number of positive pairs that are assumed to be similar)~\cite{Bhaskarrao2025}. Several approaches have been proposed to address this limitation. MoCo (Momentum Contrast) maintains a large and consistent set of negative samples by storing latent representations from previous batches, which avoids the need to recompute all negatives at each iteration~\cite{moco}. It uses two encoders, a query encoder, updated via backpropagation based on the loss gradient, and a momentum encoder, updated as the exponential moving average of the query encoder's parameters. Positive views are split so one passes through the momentum (teacher) and the other the query (student) encoder. Negative views are encoded only by the momentum encoder and stored for future use, providing a large set of negatives without the need for large batch sizes. SimSiam (Simple Siamese Representation Learning) eliminates the need for negative pairs altogether by introducing a separate projector network that adjusts the latent embedding of one of the views in a positive pair~\cite{simsiam}. This is sufficient to prevent the latent representations from collapsing, which improves computational efficiency as only positive pairs are needed in each training batch. 

Cluster-level SSL assigns multiple positive views to the same semantic group rather than matching individual instances, which enables the models to learn more generalisable representations. In DeepCluster~\cite{deepcluster}, images are encoded into latent representations that are grouped using offline $k$-means clustering. The clusters are subsequently used as pseudolabels to train the model similarly to supervised workflows. This iterative process continuously refines latent representations to follow a semantic structure without the need for human annotations. SwAV (Swapping Assignments between Views)~\cite{swav} also adopts a clustering-based approach, but instead of performing explicit offline clustering, it learns a set of prototypes that act as cluster centres and assigns latent representations to them during training~\cite{swav}. SwAV further adopts a multi-crop strategy, where each image is used to generate two large, global views and several smaller local views, where semantic consistency between these views is encouraged during the training. Each multi-crop view is processed by a shared encoder, after which their representations are assigned to clusters using the online Sinkhorn-Knopp algorithm~\cite{Sinkhorn1967ConcerningNM}. This ensures that the models are consistent in how they group different views that originated from the same image. While this generally makes results more robust to the natural variability that exists within each semantic group, the multi-crop strategy can confuse the model if semantically different content that exists within the same image gets locally isolated during cropping, and subsequently associated with the cluster. In \cite{dense-dino}, the authors demonstrated that semantic consistency can be improved by ensuring local crops partially overlap, minimising the risk of generating semantic representations from conflicting inputs. 

Self-distillation methods use a teacher-student framework, where the teacher model produces stable latent representations, and the student model is trained to align its outputs with the teacher's. Typically, the teacher's parameters are updated using an exponential moving average (EMA) of the student’s, ensuring stable and consistent targets during training. A key distinction from the teacher-student relationship in MoCo v2 is that the student in MoCo v2 doesn't attempt to match the teacher's embedding, instead it is just used as a stable reference that ensures positive views are more similar than negative views within the contrastive loss framework. Self-distillation methods such as BYOL (Bootstrap Your Own Latent)~\cite{grill2020byol} and DINO (Self-Distillation with No Labels)~\cite{dino} attempt to maximise the absolute positive pair alignment to learn representations. BYOL takes two views derived from the same image, and uses separate teacher and student models to encode latent representations. DINO combines self-distillation with clustering-like behaviours that group the latent representations of multi-crop views of the same image into the same prototype cluster, as in SWaV, and further leverages ViT architectures for the encoder model. ViT models leverage self-attention mechanisms that act on their inputs. These model dependencies across the entire positive multi-crop sequence extracted from an image. Unlike convolutional models, which expand their receptive fields in stages by stacking layers of local receptive fields, ViTs can integrate global contextual information from the very first layer, offering an alternative approach to encoding latent representations~\cite{khan2022transformers}. ViT models trained on large, curated datasets have demonstrated strong cross domain generalisation without the need for dataset-specific optimisation~\cite{dino,dinov2}.

The discriminative SSL methods described so far represent the current state-of-the-art. The SSL strategies apply during training, where once trained, only the trained encoders are needed to generate latent representations for interpretation of new images. Location metadata has been used to enhance latent representations in terrestrial scenes by partially overlapping local views as augmented pairs to capture spatial consistency using ViTs~\cite{dense-dino}. In VADeR (View-Agnostic Dense Representation)~\cite{pinheiro2020} the consistency of latent representations was maximised by providing partially overlapping patches as positive pairs for augmentation using CNNs. A potential limitation for use with seafloor images is that the SSL methods described up to this point are that they can only leverage patterns that exist within the footprint of a single image frame (1-10\,m~\cite{williams}). However, seafloor habitats and substrates vary over far greater spatial scales (tens to hundreds of metres) even in complex environments~\cite{zelada}. Our previous work has demonstrated that the quality of latent representations encoded using SimCLR was enhanced using location metadata by selecting positive pairs from images taken from physically nearby locations (GeoCLR), allowing models to capture the natural variability within semantic grouping more effectively than augmentations generated from a single image~\cite{yamada-geoclr}. Enforcing similarity across positive pairs formed from nearby images improved downstream classification by 7.5\% on a seafloor image dataset consisting of 87,000 images. However, a gap remains in understanding if the advantages seen in GeoCLR, i.e., a location-regularised modification of SimCLR, generalise across different state-of-the-art SSL techniques.

\section{Method}

\begin{figure}
    \centering
    \includegraphics[width=0.99\linewidth]{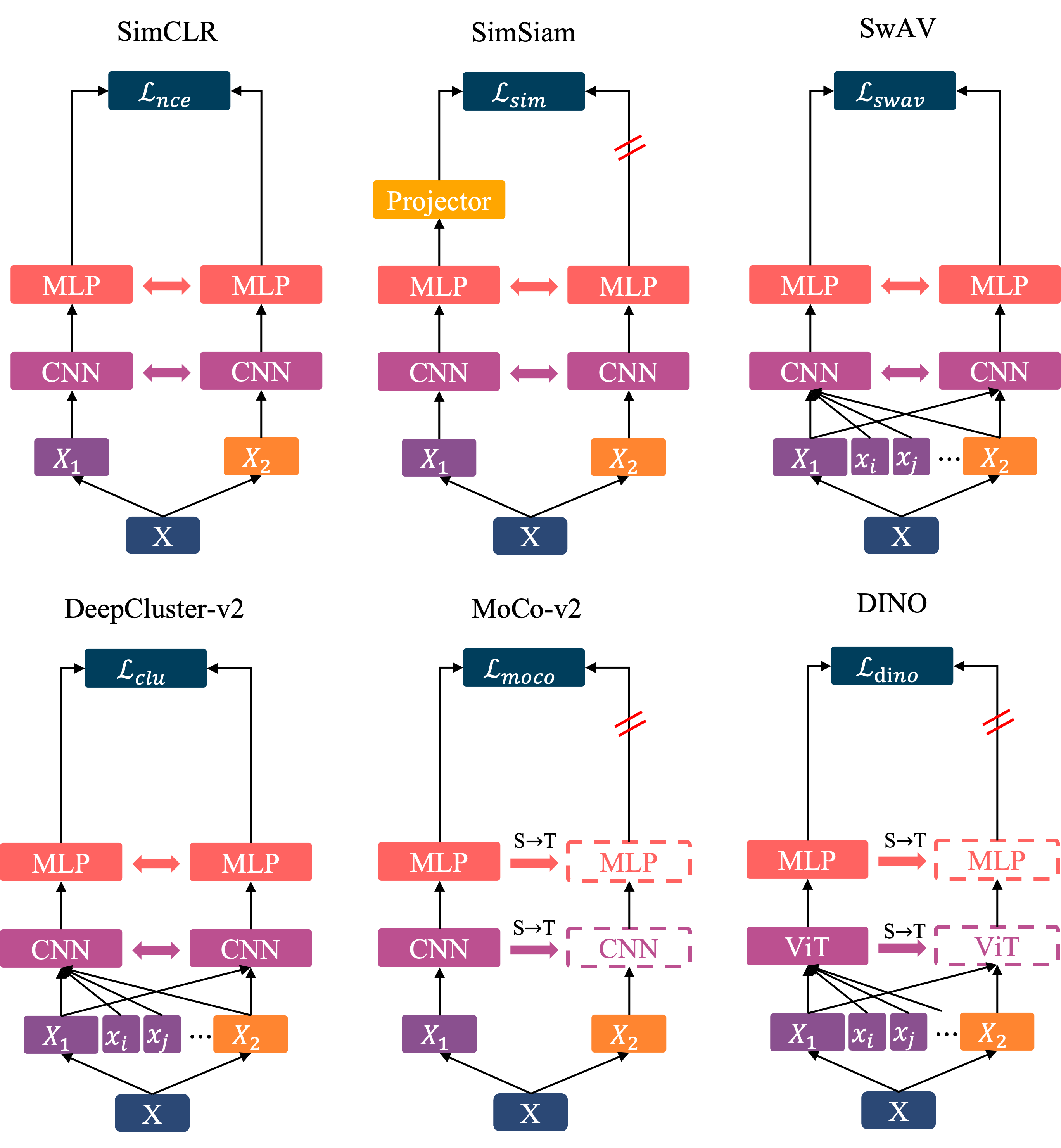}
    \caption{Discriminative SSL strategies investigated for location-regularisation. These represent instance-level contrastive (SimCLR, MoCo-v2) and non-contrastive (SimSiam) learning. SwAV and DeepCluster-v2 represent cluster-level SSL with CNNs. DINO represents self-distillation SSL with a ViT. Projector networks adjust encoder latent embeddings, and Multilayer Perceptron (MLP) heads manage dimensionality.}
    \label{fig:methods}
\end{figure}

Fig.~\ref{fig:methods} shows various SSL strategies in which we investigate location-regularisation in this study. Table~\ref{tab:ssl_comparison} summarises the main characteristics of each approach, where all methods learn latent embeddings by assuming similarity in positive views. However, if the views are too similar the latent representations become nearly identical regardless of the weights of in the CNN/ViT, offering limited signal to learn discriminative features. Conversely, if the views differ too much, the models degrade as they force semantically dissimilar inputs to nearby regions of the latent representation space. Location-regularisation forms positive views using separate images taken from nearby locations, which enforces the following assumption~\cite{yamada-lga}:\\

\noindent \textbf{Proximity assumption}: \textit{Locations that are physically close to each other are more likely to have similar seafloor characteristics than those that are far apart.}\\

\noindent During seafloor imaging surveys, robots move slowly (0.1–1.0\,m/s), following complex terrains at a low altitude while taking images at regular intervals (1-5\,s). This ensures overlap for 3D reconstruction while limiting the power consumption of their strobes~\cite{MattJoRO}. Although some images inevitably fall within transition regions between different substrate and habitat types, the vast majority of nearby images represent the same semantic group. Defining the parameter $r_{loc}$, we can control the maximum distance within which positive views can be sampled as follows:
\begin{equation}
    \sqrt{(N_i-N_j)^2+(E_i-E_j)^2} < r_{loc}, \text{where }i\neq j 
    \label{eq:distance}
\end{equation}
\noindent where $N$ and $E$ are northings and eastings coordinates, of images $i$ and $j$, respectively. Instead of generating $X_1$ and $X_2$ in Fig.~\ref{fig:methods} from a single image, we randomly select images that satisfy Eq.\ref{eq:distance}. The method's sensitivities are where:
\begin{itemize}
    \item $r_{loc}$ is too small, where no images satisfy. Here, the method generates augmented views of the same image (i.e., $i=j$ is enforced), which generates identical conditions to the base SSL methods that don't use location metadata.
    \item $r_{loc}$ is too large, where positive views are formed from images that represent semantically different seafloors scenes. Forcing dissimilar views to nearby latent representations confuses learning and causes model degradation. However, prior work has demonstrated limited impact on performance for an order of magnitude range in $r_{loc}$ due to the large size of habitats~\cite{cailei2025}.
\end{itemize}

\noindent For the positive pairs used in instance-level SSL (i.e., SimCLR, SimSiam and MoCo-v2), two global views are sampled from different images that satisfy Eq.~\ref{eq:distance}. For positive multi-crops used in cluster-level SSL (i.e., SwAV, DeepCluster-v2 and DINO), the two global views are sampled from separate images, and the subset of local views is sampled from one of the images to enforce consistency across the images at the level of clusters.

\begin{table*}[ht]
\centering
\renewcommand{\arraystretch}{1.5}
\caption{Comparison of state-of-the-art SSL strategies. These are adapted to implement location regularisation in this study.}
\begin{tabular}{lccccccc}
\hline
\textbf{Method} & \textbf{Type}  & \textbf{Negative Samples} & \textbf{Teacher-Student} & \textbf{Views} & \textbf{Backbone} & \textbf{Collapse Prevention} \\
\hline
SimCLR        & Instance-level     & Yes & No & 2 Global       & CNN        & Contrastive Learning\\
SimSiam       & Instance-level     & No & No & 2 Global       & CNN        & Asymmetry \\
MoCo-v2       & Instance-level     & Yes & Yes & 2 Global       & CNN        & Contrastive Learning + Asymmetry\\
SwAV          & Cluster-level      & No & No & Multi-crop     & CNN        & Online Prototypes \\
DeepCluster-v2& Cluster-level      & No & No & Multi-crop     & CNN        & Episodic $k$-means \\
DINO          & Self-distillation  & No & Yes & Multi-crop     & ViT        & Online Prototypes + Asymmetry \\
\hline
\end{tabular}
\label{tab:ssl_comparison}
\end{table*}

\subsubsection{SimCLR}
\begin{figure}
    \centering
    \includegraphics[width=0.8\linewidth]{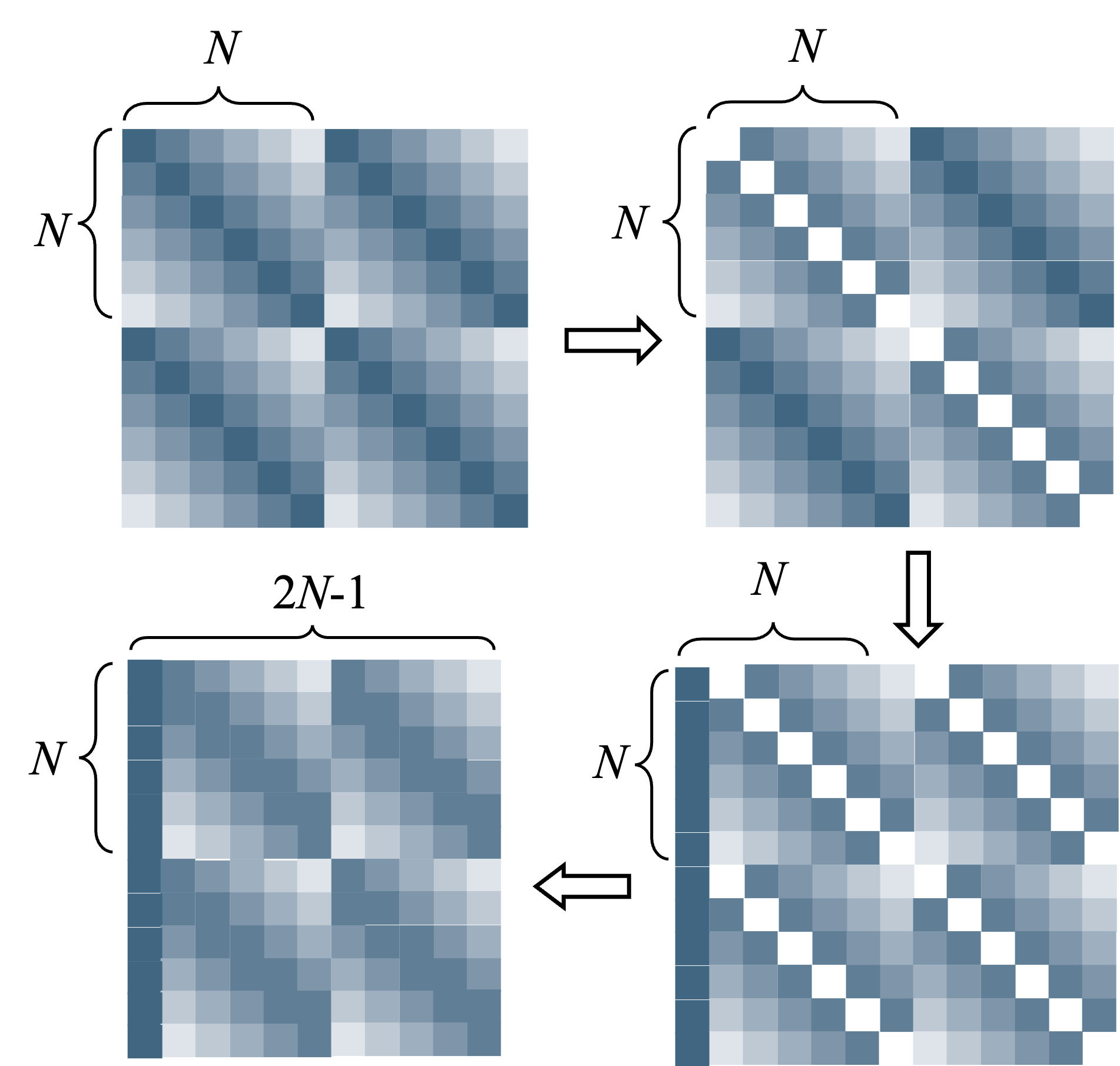}
    \caption{Constructing the SimCLR similarity matrix for a batch of $N$ images. Two augmented views are generated per image (i.e., $2N$ samples, Eq.~\ref{eq:sim_arrange}). Cosine similarities are computed between all pairs, forming a $2N \times 2N$ matrix with positive pairs on the diagonals of the off-diagonal blocks. The matrix is reformatted to $2N \times (2N-1)$ by moving positive pairs to the start of their row, followed sequentially by negative pairs.}
    \label{fig:infoNCE}
\end{figure}

SimCLR generates batches of positive and negative pairs by sampling $N$ original images, and augmenting them twice each to produce $2N$ views per batch:
\begin{equation}
[x_1^{(1)},\ x_2^{(1)},\ \dots,\ x_N^{(1)},\ x_1^{(2)},\ x_2^{(2)},\ \dots,\ x_N^{(2)}]
\label{eq:sim_arrange}
\end{equation}
where $x_i^{(1)}$ and $x_i^{(2)}$ correspond to the positive pair of views augmented from $i$th image. All other combinations of views form negative pairs. Optimising the contrastive loss function pulls the latent representations of views in each positive pair closer together and pushes those in each negative pair apart. SimCLR computes a $2N\times2N$ matrix of the cosine similarities between every pair of augmented samples (see top-left of Fig.~\ref{fig:infoNCE}). The top-left $N \times N$ block corresponds to the first set of augmentations and the bottom-right $N \times N$ block corresponds to the second. Since indices $i$ and $i+N$ represent a positive pair, the diagonals in the top-right and bottom-left $N \times N$ blocks contain positive-pair cosine similarities. To compute the loss, the diagonal similarity scores are removed (Fig.~\ref{fig:infoNCE}, top-right), and the positive pair similarities are moved to the start of each row (Fig.~\ref{fig:infoNCE}, bottom-right). The matrix is then reshaped so that each row starts with the positive pair's similarity, followed by all negative pair similarities (Fig.~\ref{fig:infoNCE}, bottom-left).

The matrix contains raw similarity scores that are used to compute the cross-entropy loss:
\begin{equation}
\mathcal{L}_i = -\log \frac{\exp\left(\mathrm{sim}\left(z_i, z_{i}^{+}\right) / \tau\right)}{\sum_{j=1}^{2N} \mathbbm{1}_{[j \neq i]} \exp\left(\mathrm{sim}\left(z_i, z_j\right) / \tau\right)}
\label{eq:single_nce}
\end{equation}

\noindent where $z_i$ and $z_i^+$ are latent representations of positive pairs. $sim(\cdot)$ computes the cosine similarity, and the parameter $\tau$ controls how sharply the optimisation distinguishes between positive and negative pairs. The total loss $L_{nce}$ (noise contrastive estimation) is expressed as:
\begin{equation}
\mathcal{L}_{nce} = \frac{1}{2N} \sum_{i=1}^{2N} \mathcal{L}_i
\label{eq:total_nce}
\end{equation}

\noindent Minimising this loss encourages the encoder to generate latent representations that capture the similarity of positive pairs. 

\subsubsection{SimSiam}

In SimCLR, representational collapse is prevented by contrasting positive pairs against large numbers of negative examples. Without this contrastive pressure, the encoder could learn the trivial solution of mapping all views to a similar latent vector, ignoring any pixel-level variations. SimSiam eliminates the need for negative pairs by introducing asymmetry. After encoding two augmented views $X_1$ and $X_2$ into latent vectors $z_1$ and $z_2$, a learnable projector network $p(\cdot)$ is introduced to modify one of the vectors to $p_1=p(z_1)$ or $p_2=p(z_2)$, as shown in Fig.~\ref{fig:methods}. The loss function then measures the cosine similarity between each projector output and the opposing pair's encoder output using Eq.~\ref{eq:loss_simsiam}:

\begin{equation}
\mathcal{L}_{sim} = - \frac{1}{2} \left( \mathrm{sim}(p_1, z_2) + \mathrm{sim}(p_2, z_1) \right)
\label{eq:loss_simsiam}
\end{equation}

\noindent This encourages the representations of the two views to align. When comparing the projector output $p_i$ to the encoder output $z_j$ (where $i\neq j$), the encoder output is fixed while the projector and the upstream encoder are optimised together to make $p_i$ match the fixed target. This asymmetric update rule prevents the network from collapsing to a constant output. 

\subsubsection{MoCo v2}
MoCo v2 is an instance-level contrastive SSL that reduces computational costs relative to SimCLR by storing and reusing negative representations from earlier training iterations. It uses a teacher-student framework, where one view is processed by a student encoder and the other is processed by a teacher encoder. The teacher's parameters are updated using a moving average of the student encoder’s parameters:

\begin{equation}
\theta_t = m \cdot \theta_t + (1 - m) \cdot \theta_s
\label{eq:ema}
\end{equation}
\noindent where $\theta_t$ and $\theta_s$ denote the parameters of the teacher and student models, respectively, and $m$ is the momentum coefficient. The following loss function is used:
\begin{equation}
\mathcal{L}_{moco} = -\log \frac{\exp\left(\mathrm{sim}(q, k^+)/\tau\right)}{\exp\left(\mathrm{sim}(q, k^+)/\tau\right) + \sum_{k^-} \exp\left(\mathrm{sim}(q, k^-)/\tau\right)}
\label{eq:moco_loss}
\end{equation}
\noindent where $q$ denotes the student encoder embedding, $k^+$ denotes the teacher encoder embedding, and $k^-$ denotes the negative embeddings sampled from earlier training iterations. $sim(\cdot)$ and $\tau$ are used in the same way as in SimCLR.

\subsubsection{SwAV}
A cluster-level SSL that uses a multi-crop strategy. Each image is randomly cropped into two global views ($X_1, X_2$) and multiple local views ($x_i$ for $i=1,2,\dots,N$), where each view is independently augmented. Global views typically have $\sim5\times$ more pixels than local ones. The small size of local views helps enrich learning while minimising the computational overhead. After computing the latent representations, SwAV uses the Sinkhorn-Knopp algorithm to compute the probability of each global view belonging to a set of prototype cluster centres. The cross-entropy loss is computed between each global view's cluster probability distribution and those of the remaining global and local views. The encoder's optimisation minimises the difference between each multi-crop view's cluster assignment for semantic alignment. 

\begin{equation}
\mathcal{L}_{swav} =  \frac{1}{2N} \sum_{i=1}^{N} \sum_{v \in \mathcal{V}_g} \sum_{\substack{v' \in \mathcal{V}_g \cup \mathcal{V}_l \\ v' \neq v}} - q_{i}^{(v)^\top} \log p_{i}^{(v' \to v)}
\label{eq:swav_loss}
\end{equation}

\noindent where $N$ is the batch size. $\mathcal{V}_g$ consists of two global views. $\mathcal{V}_l$ contains several local views. $q_{i}^{(v)}$ denotes the Sinkhorn assignment from global views and $p_{i}^{(v' \to v)}$ is the cluster assignment predicted for the global or local view $v'$.

\subsubsection{DeepCluster-v2}
This multi-crop cluster-level SSL method trains the model using discrete cluster assignments obtained via offline $k$-means. The loss function is defined as:
\begin{equation}
\mathcal{L}_{clu} = \frac{1}{N} \sum_{i=1}^{N} \mathrm{CE}\left(p_i, y_i\right) = - \frac{1}{N} \sum_{i=1}^{N} \sum_{k=1}^{K} \mathbbm{1}_{[y_i = k]} \log p_i^k
\end{equation}
where $N$ is the batch size. $K$ is the predefined number of clusters, and $y_i$ in $\{1, \dots, K\}$ is the discrete cluster label assigned to sample $i$ via $k$-means. The model predicts a probability distribution $p_i = [p_i^1, \dots, p_i^K]$, where $p_i^k$ is the predicted probability of sample $i$ belonging to cluster $k$. $\mathrm{CE}(\cdot)$ is the cross-entropy loss. Unlike SwAV, DeepCluster-v2 performs clustering at the end of each training iteration using the latent representations of all samples in the dataset. Although this has a high memory and computational cost compared to the online clustering in SwAV, it enables more accurate and globally consistent partitioning of the latent space even during early stages of training.

\subsubsection{DINO}
DINO combines self-distillation without labels, leveraging momentum encoders and cluster-level multi-crop SSL strategies to train ViT models, and has demonstrated strong generalisation across a wide range of tasks~\cite{dino}. The training loss is computed as:
\begin{equation}
\mathcal{L}_{dino} = - \frac{1}{|\mathcal{V}_s|} \sum_{v_s \in \mathcal{V}_s} \frac{1}{|\mathcal{V}_t|} \sum_{v_t \in \mathcal{V}_t} \sum_{k=1}^{K} P_t^{v_t, k} \log P_s^{v_s, k}
\label{eq:dino_loss}
\end{equation}
Each crop per view is processed by the student and teacher networks to produce latent vectors $v_s$ and $v_t$. These are converted to prototype cluster probabilities distributions over $K$ prototypes. Here, $P_s^{v_s, k}$ and $P_t^{v_t, k}$ represent how strongly the student and teacher associate with the cluster prototype $k$. $\mathcal{V}_s$ and $\mathcal{V}_t$ represent the set of student and teacher views, respectively, where $|\cdot|$ give the total number of views for output normalisation. The self-distillation framework here differs from MoCo-v2's teacher-student as both student and teacher outputs are directly used to compute the loss~\cite{grill2020byol}.

\section{Results}

\subsection{Dataset Description}
\begin{table*}[]
    \centering
    \caption{Dataset description. DM, CB and SH datasets are collected from three different locations. The values indicated for depth, resolution and patch interval are averages for each dataset. The labels are human-verified references used for downstream classifier training and performance validation. These are hosted on the marine image repository Squidle+ (\url{https://squidle.org/})}
    \label{tab:dataset-description}
    \renewcommand{\arraystretch}{1.5}
    \begin{tabular}{l|ccc}
     & Darwin Mounds (DM) &  Cawsand Bay (CB) & Southern Hydrate Ridge (SH) \\
    \hline
    Habitat type & Cold-water Coral & Mixed sediment and rock & Gas hydrates and infrastructure\\
    Number of classes & 4 & 3 & 6\\    
    Location & $59.814^\circ\mathrm{N}$, $7.358^\circ\mathrm{W}$ & $50.338^\circ\mathrm{N}$, $4.178^\circ\mathrm{W}$ &
    $44.571^\circ\mathrm{N}$, $125.149^\circ\mathrm{W}$ \\
    Seafloor depth, m & 970 & 12 & 775\\
    \hline
    AUV (Camera system) &  Autosub6000 (BioCam) & Smarty200 (Recon) & AE2000f (Seaxerocks)\\
    Image resolution, mm  & 3 & 0.4 &  5.3 \\    
    Image patch interval, m & 3.6 & 0.5 & 1.8\\
    Labeled image patches & 18,575 & 48,230 & 27,000
    \end{tabular}
\end{table*}

\begin{figure*}
\centering
    \includegraphics[width=.8\linewidth]{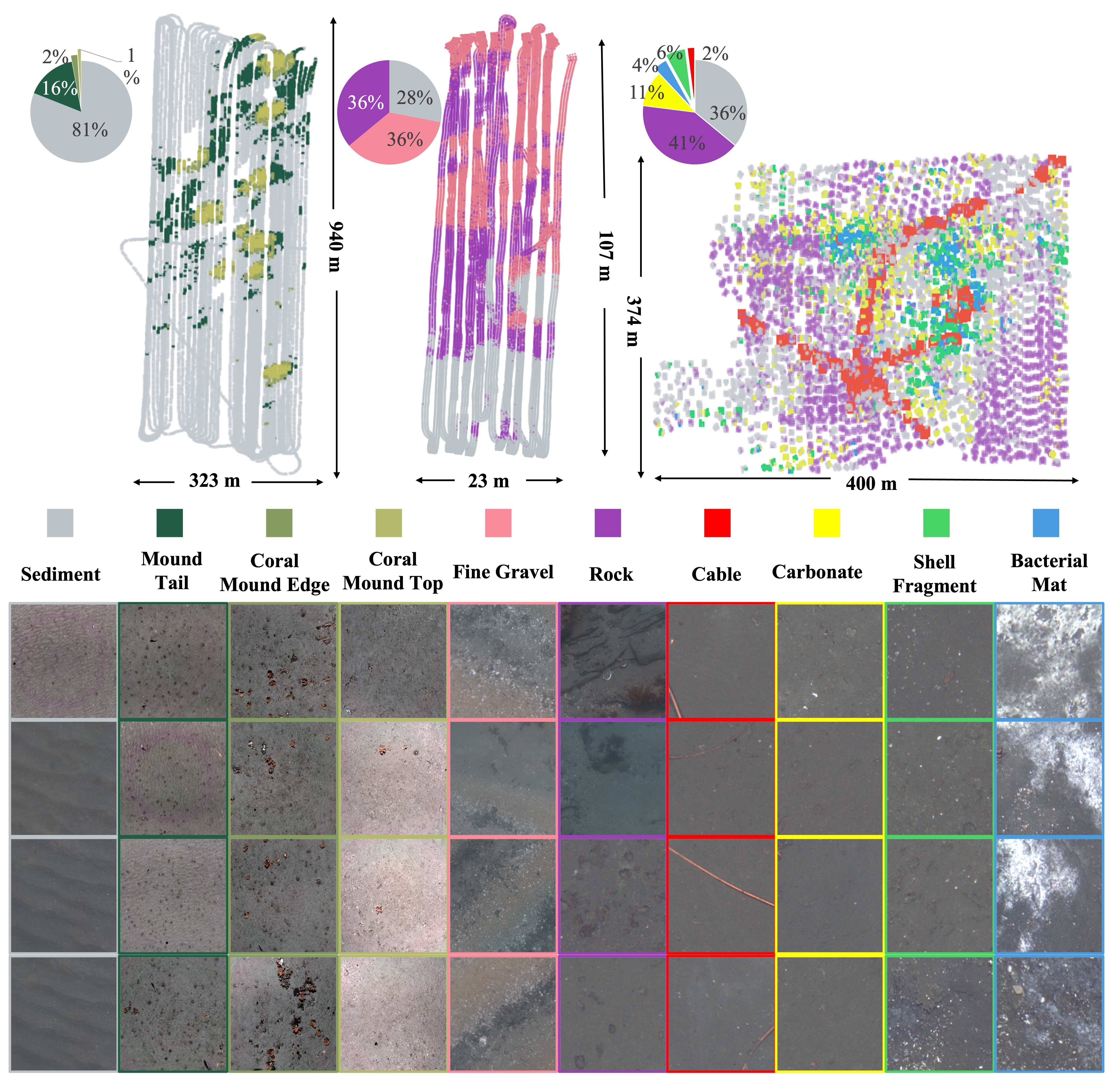}
    \caption{The datasets used in this study consist of more than 90k labelled images. These were obtained at various sites using different AUVs and imaging systems. They include diverse habitats, imbalanced class distributions and different image sizes and resolutions. The top row shows the class distribution and proportions for the DM (left), CB (centre) and SH (right) datasets. The bottom shows example images for each class from across the sites.}
    \label{fig:dataset-overview}
\end{figure*}

Three diverse seafloor image datasets consisting of 90k AUV images with human-validated labels are used to characterise the performance of the SSL methods. These were obtained from various locations and depths, representing different habitats and data acquisition conditions (hardware, altitude). Fig.~\ref{fig:dataset-overview} gives examples of each visual class, their spatial distribution and proportion at each site. Table~\ref{tab:dataset-description} gives an overview of each dataset.

The Darwin Mounds (DM) dataset was collected in 2019 from the Darwin Mounds Special Area of Conservation (SAC), located 160\,km northwest of Cape Wrath, UK, at a depth of 970\,m. The National Oceanography Centre's Autosub6000 AUV was equipped with the University of Southampton's BioCam imaging system. The dataset consists of 18,575 seafloor images gathered from a target altitude of 5\,m. These were automatically classified following method outlined in~\cite{yamada-geoclr}, and all outputs were manually checked and refined by human experts to generate reference class labels. These consist of four imbalanced classes: Sediment (81\%), Mound Tail (16\%), Coral Mound Edge (2\%), and Coral Mound Top (1\%). A random subset of 14,860 images (without labels) were used to train the SSL feature extractors. The corresponding labels were used to train a downstream support-vector machine (SVM) to delineate the latent space. The remaining images and labels were used to assess the performance. 

The Cawsand Bay (CB) dataset was collected in 2024 off Plymouth, UK, using the University of Southampton's Smarty200 AUV equipped with a Voyis Recon imaging system. The dataset consists of 48,230 images gathered from 1.5\,m altitude. Reference labels were generated following the same process as DM, and consisted of three balanced classes: Sediment (28\%), Fine Gravel (36\%), and Rock (36\%). SSL feature extractors were trained on all images, and a downstream SVM was trained using a random subset of 19,292 labels. The remaining images and labels were used to assess the performance. 

The Southern Hydrate Ridge (SH) dataset was collected in 2018 off Oregon, USA, using the University of Tokyo's AE2000f AUV equipped with the SeaXerocks imaging system. The dataset consists of 27,000 images gathered from a target altitude of 6\,m. References labels were generated by human-experts~\cite{yamada-lga}, consisting of six imbalanced classes: Sediment (36\%), Rock (41\%), Carbonate (11\%), Bacterial Mat (4\%), Shell Fragment (6\%), and Cable Infrastructure (2\%). SSL feature extractors were trained on all images, where the downstream SVM was trained using a random subset of 10,800 labels. The remaining images and labels were used to assess the performance. 

While the numbers of images used for SSL and SVM training vary between dataset, our prior work has shown that performance converges beyond 200 images per class~\cite{yamada-pami, grant}. The number of training examples satisfies this condition for all classes used in this study. 

\subsection{Experimental Setup}

We evaluate the accuracy of downstream classifiers across three datasets using latent representations extracted by six SSL methods: SimCLR, SimSiam, MoCo-v2, SwAV, DeepCluster-v2 and DINO, along with their location-regularised counter-parts: GeoCLR, GeoSimSiam, GeoMoCo-v2, GeoSwAV, GeoDeepCluster\_v2 and GeoDINO. Location-regularisation was implemented with $r_{loc}$ in Eq.~\ref{eq:distance} set to 4.0, 0.5, and 2.0\,m for DM, CB, and SH, respectively. These values round the image intervals to the nearest 0.5\,m for each dataset, limiting positive pairs to immediately neighbouring patches. 

All CNN-based SSL models used a ResNet18 backbone, which was fully trained using the corresponding target dataset. For the ViT-based model, we fine-tuned an ImageNet-1k pre-trained model by freezing different numbers of transformer blocks (2, 4, 8, and 12) to leverage its zero-shot transfer capabilities. These configurations are denoted as FT2B, FT4B, FT8B, and FT12B, respectively, and their location-regularised counterparts are denoted as GeoFT2B, GeoFT4B, GeoFT8B, and GeoFT12B. All models were trained to converge on the training loss.

Models were evaluated with default latent dimensions (512 for CNNs, 768 for ViTs) and reduced dimensions (128 for both) using PCA, following prior methods for CNNs~\cite{cnn-pca} and ViTs~\cite{vit-pca}. Low-dimensional representations can reduce overfitting and improve generalisation, especially with limited data. SSL performance was compared to equivalent zero-shot CNN~\cite{paszke2019pytorch} and ViT~\cite{dino} models, using publicly released parameters (with the DINO model pre-trained on ImageNet-1k). For these comparisons, only the downstream classifiers were trained on the target datasets. Performance was evaluated with macro-F1 scores for robustness to class imbalance.

\subsection{Discussion}
\begin{table*}[ht]
\caption{CNN model-based downstream classification F1-scores for each dataset using latent representations extracted by the pre-trained model and dataset-optimised models using each of the SSL methods together with their location-regularised counterparts. The top half of the table uses ResNet18's original 512-dimension latent representations and the bottom half uses 128-dimension latent representations obtained through dimensionality reduction. The highest scoring approach for each dataset and network configuration is in bold, with the second highest in italics.}
\label{tab:cnn-results}
\centering
\footnotesize
\renewcommand{\arraystretch}{1.5}
\begin{tabular}{c|c|ccccc|ccccc}
\hline
Dataset & \makecell{Pre-trained\\ResNet18} & 
\makecell{SimCLR} & \makecell{SimSiam} & \makecell{MoCo-v2} & \makecell{SwAV} & \makecell{DeepCluster\\v2} & 
\makecell{GeoCLR} & \makecell{Geo\\SimSiam} & \makecell{Geo\\MoCo-v2} & \makecell{Geo\\SwAV} & \makecell{Geo\\DeepCluster-v2} \\

\hline
DM & 
0.767 & 
0.768 & 0.753 & 0.755 & 0.744 & 0.787 & 
\textbf{0.814} & 0.798 & \textit{0.802} & 0.791 & 0.798  \\
CB & 
0.824 & 
0.831 & 0.810 & 0.817 & 0.831 & 0.834 & 
\textbf{0.863} & 0.838 & \textit{0.857} & 0.835 & 0.845    \\
SH & 
0.611 & 
0.573 & 0.527 & 0.573 & 0.586 & 0.552 & 
\textbf{0.647} & \textit{0.631} & 0.616 & 0.597 & 0.590  \\
\hline
Avg & \makecell{0.734 \\$\pm$ 0.107} & \makecell{0.724 \\$\pm$ 0.132} & \makecell{0.697 \\$\pm$ 0.145} & \makecell{0.715 \\$\pm$ 0.132} & \makecell{0.720 \\$\pm$ 0.128} & \makecell{0.724 \\$\pm$ 0.143} & \makecell{\textbf{0.775} \\\textbf{$\pm$ 0.111}} & \makecell{0.756 \\$\pm$ 0.106} & \makecell{\textit{0.758} \\\textit{$\pm$ 0.121}} & \makecell{0.741 \\$\pm$ 0.120} & \makecell{0.744 \\$\pm$ 0.131} \\

\hline
\hline
DM & 
0.752 & 
0.773 & 0.761 & 0.755 & 0.731 & 0.789 & 
\textbf{0.825} & 0.796 & \textit{0.823} & 0.775 & 0.815  \\
CB & 
0.823 & 
0.829 & 0.823 & 0.815 & 0.829 & 0.833 & 
\textbf{0.862} & 0.837 & \textit{0.852} & 0.835 & 0.836   \\
SH & 
0.490 & 
0.601 & 0.571 & 0.602 & 0.531 & 0.543 & 
\textbf{0.647} & 0.590 & \textit{0.622} & 0.534 & 0.557  \\
\hline
Avg & \makecell{0.688 \\$\pm$ 0.168} & \makecell{0.734 \\$\pm$ 0.114} & \makecell{0.718 \\$\pm$ 0.127} & \makecell{0.724 \\$\pm$ 0.106} & \makecell{0.697 \\$\pm$ 0.150} & \makecell{0.722 \\$\pm$ 0.146} & \makecell{\textbf{0.778} \\\textbf{$\pm$ 0.122}} & \makecell{0.741 \\$\pm$ 0.124} & \makecell{\textit{0.766} \\\textit{$\pm$ 0.116}} & \makecell{0.715 \\$\pm$ 0.151} & \makecell{0.736 \\$\pm$ 0.144} \\

\hline

\end{tabular}
\end{table*}

\begin{table*}[ht]
\caption{ViT model-based downstream classification F1-scores for each dataset using latent representations extracted by the pre-trained model and the dataset-optimised models that were fine-tuned to different block depths, together with their location-regularised counterparts. The top half of the table uses DINO's original 768-dimension latent representation space and the bottom half uses 128-dimension latent representations obtained through dimensionality reduction. The highest scoring approach for each dataset and network configuration is in bold, with the second highest in italics.}
\label{tab:vit-results}
\centering
\footnotesize
\renewcommand{\arraystretch}{1.5}
\begin{tabular}{c|c|cccc|cccc}
\hline
Dataset & Pre-trained DINO & FT12B & FT8B & FT4B & FT2B & GeoFT12B & GeoFT8B & GeoFT4B & GeoFT2B \\
\hline
DM & \textbf{0.826} & 0.675 & 0.727 & 0.788 & 0.812 & 0.803 & 0.813 & \textbf{0.826} & \textit{0.823} \\
CB & 0.850 & 0.834 & 0.840 & 0.841 & 0.849 & 0.846 & 0.847 & \textit{0.853} & \textbf{0.855} \\
SH & \textbf{0.709} & 0.658 & 0.676 & 0.683 & 0.688 & 0.703 & 0.690 & 0.685 & \textit{0.707} \\
\hline
Avg & \makecell{\textbf{0.795} \\\textbf{$\pm$ 0.072}} & \makecell{0.722 \\$\pm$ 0.091} & \makecell{0.748 \\$\pm$ 0.082} & \makecell{0.771 \\$\pm$ 0.080} & \makecell{0.783 \\$\pm$ 0.081} & \makecell{0.784 \\$\pm$ 0.072} & \makecell{0.783 \\$\pm$ 0.079} & \makecell{\textit{0.788} \\\textit{$\pm$ 0.085}} & \makecell{\textbf{0.795} \\ \textbf{$\pm$ 0.083}} \\
\hline
\hline

DM & \textit{0.806} & 0.692 & 0.794 & 0.757 & 0.776 & \textbf{0.810} & \textit{0.806} & 0.780 & \textit{0.806} \\
CB &   0.827  & 0.833 & 0.821 & 0.832 & 0.840 & \textbf{0.844} & 0.840 & 0.834& \textit{0.841} \\
SH & 0.602   & 0.603 & 0.521 & 0.509 & 0.491 & 0.578  & \textbf{0.672} & 0.596 & \textit{0.623}\\
\hline
Avg & \makecell{0.745 \\$\pm$ 0.114} & \makecell{0.709 \\$\pm$ 0.109} & \makecell{0.712 \\$\pm$ 0.148} & \makecell{0.699 \\$\pm$ 0.163} & \makecell{0.702 \\$\pm$ 0.176} & \makecell{0.744 \\$\pm$ 0.133} & \makecell{\textbf{0.773} \\\textbf{$\pm$ 0.086}} & \makecell{0.737 \\$\pm$ 0.119} & \makecell{\textit{0.757} \\\textit{$\pm$ 0.112}} \\

\hline
\end{tabular}
\end{table*}

Table~\ref{tab:cnn-results} compares CNN model-based classification results using latent features extracted by the pre-trained model, and dataset-optimised models that used various SSL methods and their location-regularised variants. For high latent dimensions, the pre-trained ResNet18 (F1-score 0.734$\pm$0.107) outperforms all dataset-optimised models that did not use location metadata, which averaged 0.018 lower (i.e., -2.4\% relative performance).  However, location-regularisation consistently improved the score by an average of 0.021 (+2.8\%) and 0.039 (+5.4\%) compared to the pre-trained and location-unaware SSL models. The best-performing model was GeoCLR, which achieved an F1-score of 0.775$\pm$0.113,  improving by 0.041 (+5.6\%) over the pre-trained model. The enhancement from SSL and location-regularisation are greater for the low-dimensional models. Even though the pre-trained model performs poorly with an F1-scores of 0.688$\pm$0.168, the SSL models averaged 0.059 higher (+8.6\%) with location regularisation, and 0.031 higher (+4.5\%) without. Location regularisation enhanced the latent representation by 0.028 (+3.8\%) on average across the SSL methods. The best-performing low-dimensional model was GeoCLR, which achieved 0.778$\pm$0.122, improving by 0.115 (+13.1\%) over the corresponding pre-trained model. SSLs that used positive pairs outperformed the multi-crop strategy when using location-regularisation. This degradation may stem from semantic ambiguity, where local crops may be more susceptible to lacking the global context needed to capture consistent semantic patterns across images, resulting in conflicting signals during training.

\begin{figure*}[h!]
    \begin{subfigure}[t]{0.49\linewidth}
        \centering
        \includegraphics[width=0.46\linewidth]{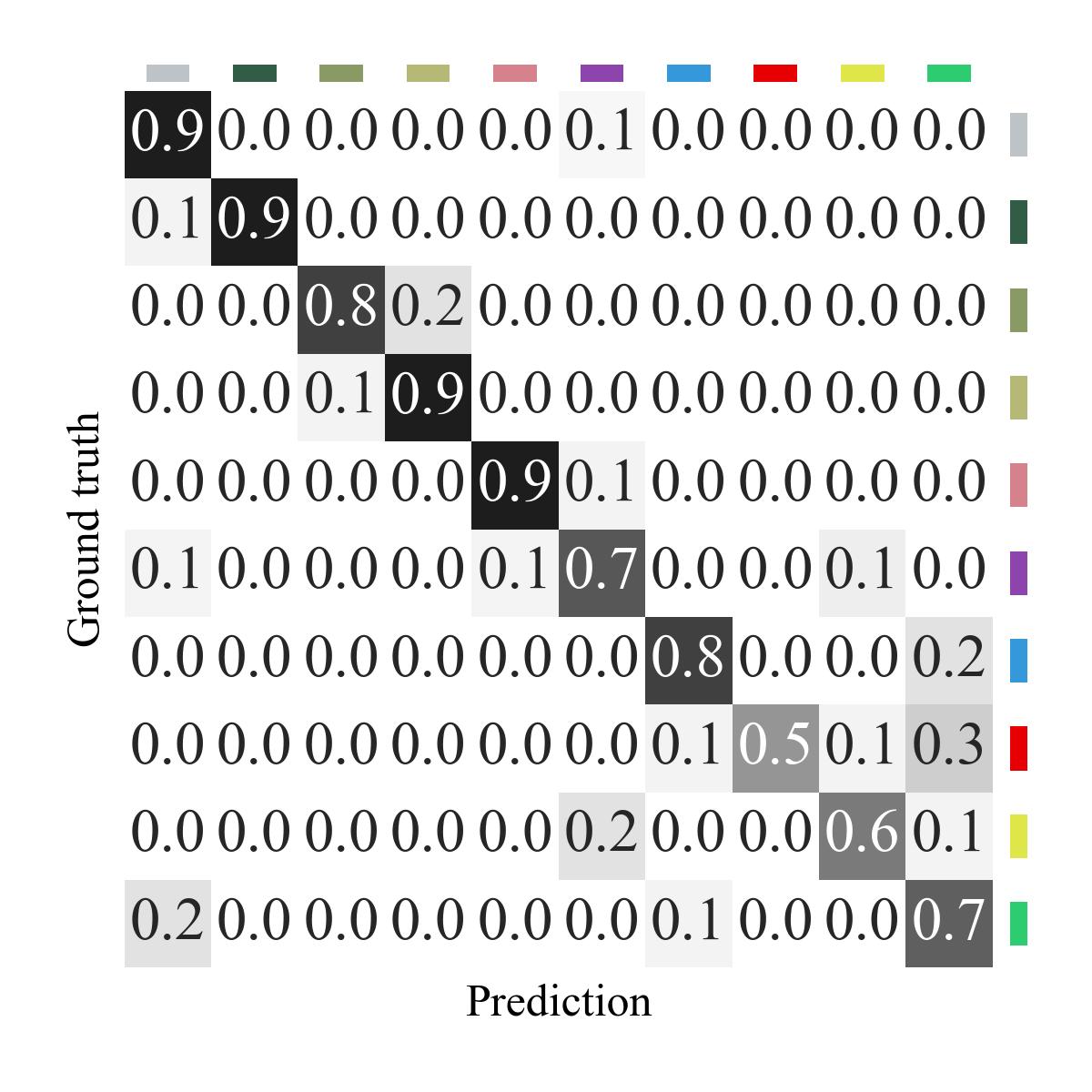}
        \includegraphics[width=0.46\linewidth]{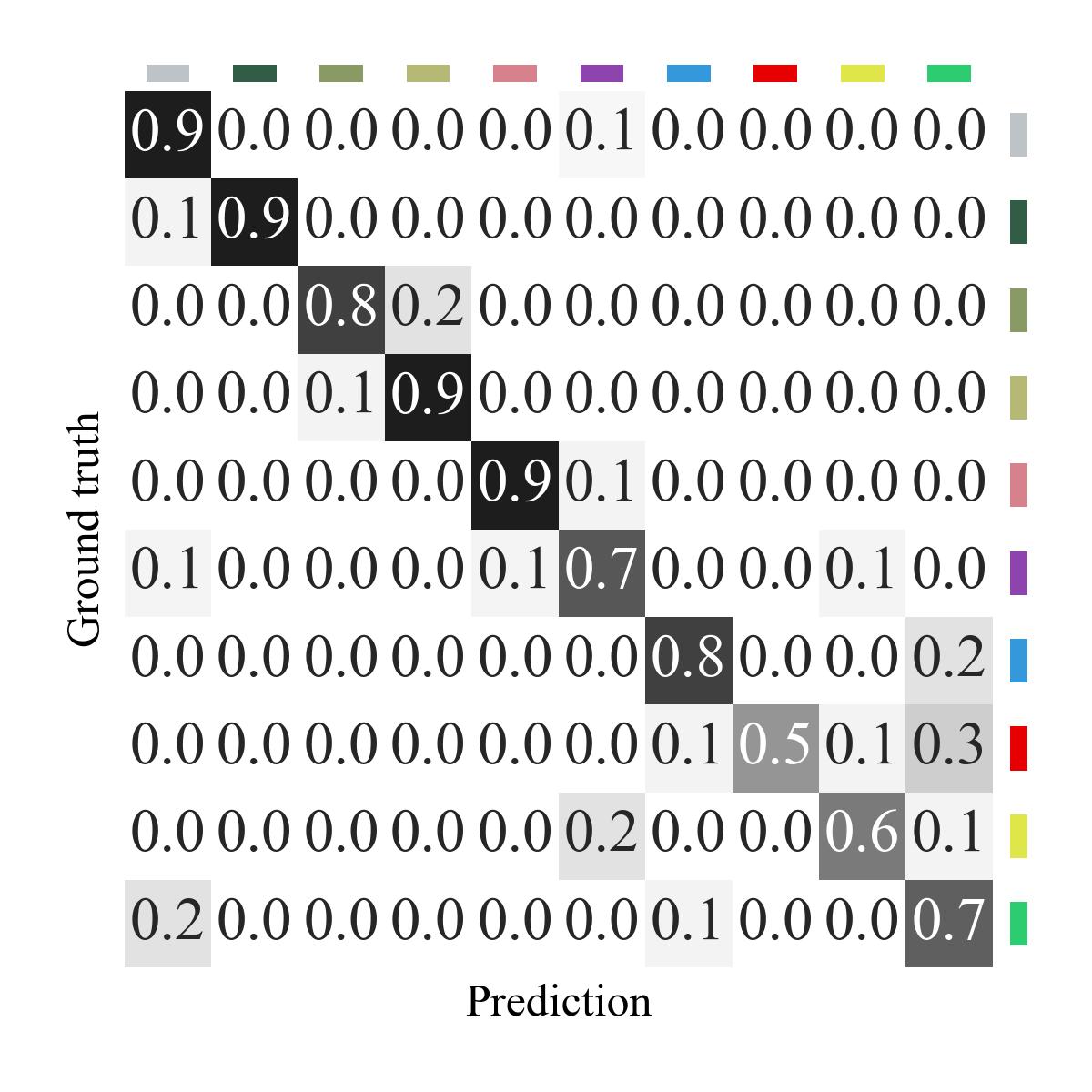}
        \caption{}
    \end{subfigure}
    \hfill
    \begin{subfigure}[t]{0.49\linewidth}
        \centering
        \includegraphics[width=0.46\linewidth]{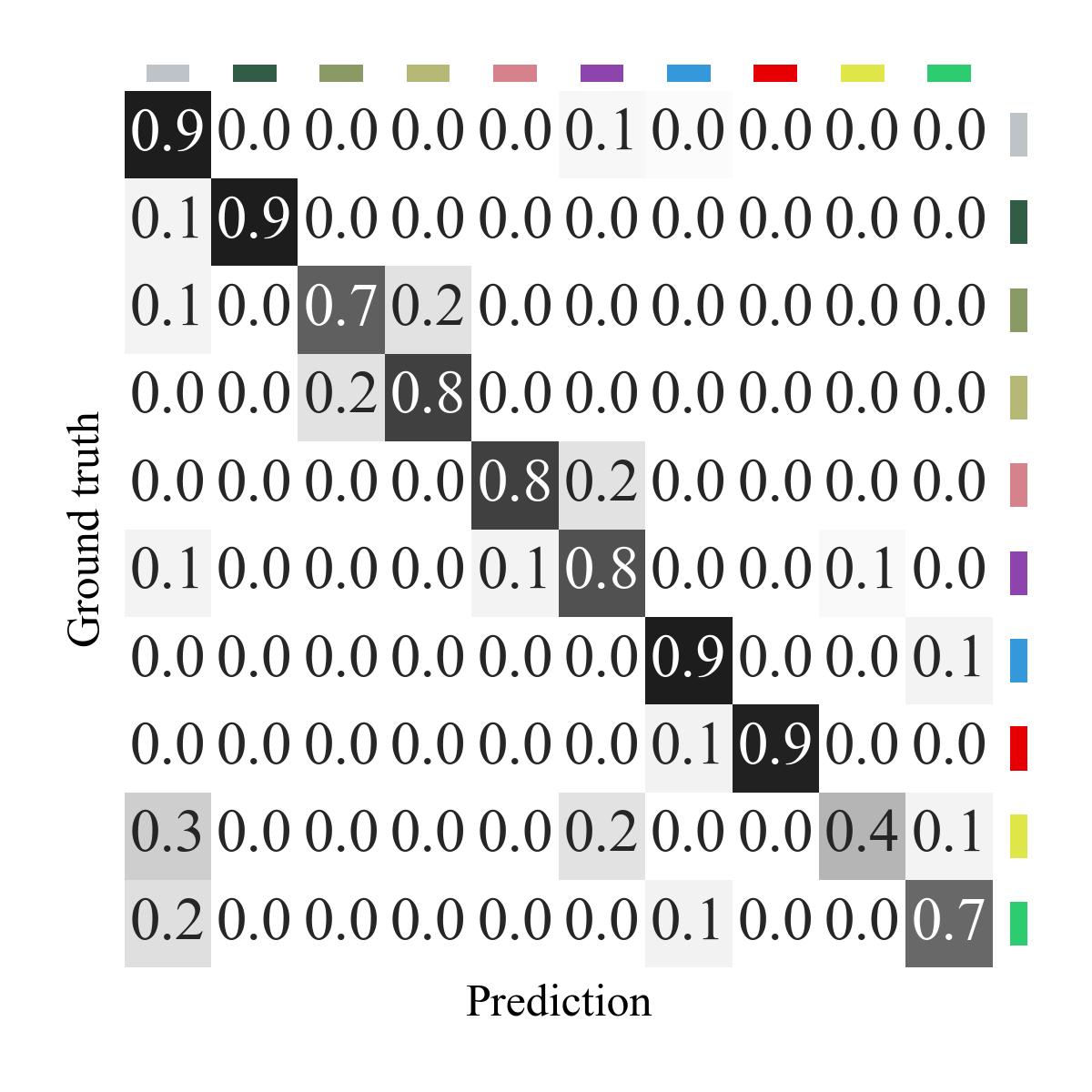}
        \includegraphics[width=0.46\linewidth]{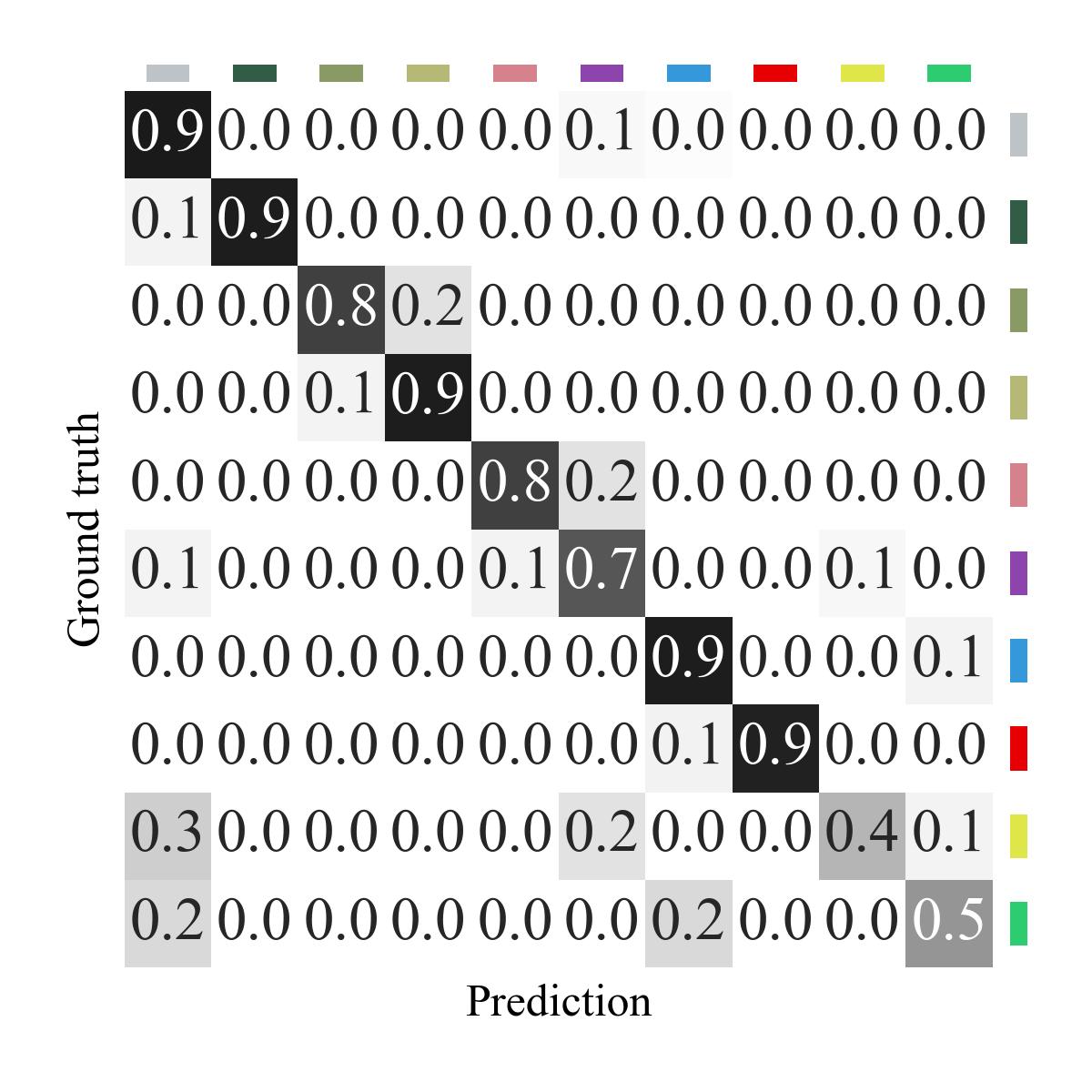}
        \caption{}
    \end{subfigure}
    \caption{Confusion matrices of the top-performing CNN (a) and ViT (b) models, where the left and right of each subfigure show the best pre-trained models and dataset-optimised SSL models, respectively. The high-dimensional pre-trained and location-regularised GeoFT12B ViT models outperformed the CNN models. The best low-dimensional model across all conditions was the GeoCLR CNNs. The colour scheme used for labelling is consistent with that shown in Fig.~\ref{fig:dataset-overview}.}
    \label{fig:heatmaps}
\end{figure*}

Table~\ref{tab:vit-results} presents the results for pre-trained ViT models, and target SSL models that were fine-tuned with different frozen transformer blocks, with and without location-regularisation. A notable difference from the trends with the pretrained CNNs is that high-dimensional models perform consistently better than the low-dimensional models across all training conditions. While location regularisation improves F1-scores over standard SSL fine-tuning, the improvements are smaller than with CNNs, achieving 0.032 (+4.19\%) and 0.047 (+6.62\%) higher F1-scores for high and low-dimensional models, respectively. Notably, however, location-regularised models did not achieve any general advantage over pre-trained models. For high-dimensionality, the shallow fine-tuned model (GeoFT2B) matched the pre-trained performance, achieving an F1-score of 0.795$\pm$0.083 compared to 0.795$\pm$0.072, whereas for low-dimensional models deeper fine-tuning was effective, with GeoFT8B outperforming the pre-trained model by 0.028 (+3.8\%). On average, however, the difference between location-regularised models and pre-trained models was marginal. 

Overall, the results show consistent performance improvements to SSL through location-regularisation for both CNNs and ViTs. They also demonstrate the efficiency of location-regularisation for the CNN, which achieved the highest F1-score for low-dimensional latent representations across all conditions. The results demonstrate the strong generalisation of high-dimensional ViTs, which achieved the highest performance across all conditions, matching the best location-regularised ViT (high-dimensional GeoFT2B) and outperforming the best CNN (low-dimensional GeoCLR). The class confusion matrices of the top-performing models are shown in Fig.~\ref{fig:heatmaps}. While trends are similar, the ViT models were better at recognising cables and infrastructure, but more likely to confuse carbonates and shell fragments with sediments compared to CNN models. 

We note that both training and inference with ViTs require greater memory and computational resources due to their larger architecture (86 million parameters), in contrast to ResNet18 (11.7 million parameters). We further note that high-dimensional latent spaces typically need larger datasets in order to train downstream classifiers and perform downstream tasks such as online clustering. These constraints need to be considered when designing any real-time robotic applications.

\section{Conclusion}

Location regularisation improves the performance of state-of-the-art self-supervised learning strategies across CNN and ViT architectures and latent-space dimensionalities. Downstream classification experiments across three diverse seafloor datasets consisting of 90k images and 10 classes demonstrated an average improvement of 4.9$\pm$4.0\% and 6.3$\pm$8.9\% in F1-score through location regularisation relative to their SSL counterparts for CNNs and ViTs, respectively. The maximum improvements across the datasets were 9.7$\pm$8.8\% for CNNs and 10.9$\pm$15.6\% for ViTs. Compared to equivalent pre-trained models, the dataset-optimised location-regularised SSLs improved CNNs by 6.9$\pm$7.8\%, with the best performing location-regularised CNN achieving a 15.4$\pm$9.8\% higher F1-score. For ViTs however, there was no enhancement of the latent representation compared to the pre-trained models (0.07$\pm$0.03\%), where the best-performing location-regularised ViT's performance was matched by its pre-trained counterpart. 

A notable difference between the CNN and ViT models is the strong generalisation of pre-trained ViTs, especially when using high-dimensional latent representations. These models achieve an F1-score of 0.795$\pm$0.072, matching the best dataset-optimised ViT (GeoFT2B), a shallow fine-tuned model with location-regularisation, which achieved similar performance (0.795$\pm$0.083). High-dimensional ViTs consistently outperformed their low-dimensional counterparts. Conversely, pre-trained CNN models showed poor generalisation. Although high-dimensional pre-trained models outperformed low-dimensional ones, dataset-specific optimisation with location regularisation consistently improved performance across all SSL methods and dimensionalities. Location-regularised models with paired positive views outperformed those with positive multi-crop strategies. Notably, low-dimensional CNNs matched or exceeded the performance of their high-dimensional counterparts, indicating high representational efficiency. The best low-dimensional model overall was GeoCLR, achieving an F1-score of 0.778$\pm$0.122.

In terms of practical considerations, although performance trends were consistent across the datasets, there was significant variability in F1 scores achieved between datasets. This can be attributed to the different habitats, class structures and image quality achieved by the various data acquisition conditions. The ViT models are significantly larger than the CNNs (86 million parameters compared to ResNet18's 11.7 million parameters), which requires greater computational resources to run and larger datasets to train. Furthermore, high-dimensional latent spaces typically need larger datasets for downstream tasks such as clustering and classification.

\section{Acknowledgment}
The research was carried out under the Innovate UK OASIS project (10110715). Darwin Mounds data was gathered during NERC CLASS (NE/R015953/1) and Oceanids BioCam cruises of the RRS Discovery (NE/P020887/1 and NE/P020739/1). We thank Dr. Veerle Huvenne for assisting in generating human-verified labels for the Darwin Mounds dataset. The Southern Hydrate Ridge data was gathered during the Schmidt Ocean Institute's \#AdaptiveRobotics campaign (FK180731). The Cawsand Bay dataset was gathered using the University of Southampton's Smarty200 AUV (EPSRC EP/V035975/1). Cailei Liang is funded by the China Scholarship Council.
\bibliographystyle{IEEEtran}
\bibliography{reference}

\end{document}